# SAR AND OPTICAL DATA FUSION BASED ON ANISOTROPIC DIFFUSION WITH PCA AND CLASSIFICATION USING PATH-BASED SVM WITH LBP

*Achala Shakya* [1]*, *Mantosh Biswas*[1], *Mahesh Pal*[2]

[1] Computer Engineering Department, National Institute of Technology, Kurukshetra, India;

[2] Civil Engineering Department, National Institute of Technology, Kurukshetra, India

*Corresponding Author: E-mail: achala_6180071@nitkkr.ac.in, shakyaachala@gmail.com

## ABSTRACT

SAR (VV and VH polarization) and optical data are widely used in image fusion to use the complimentary information of each other and to obtain the better-quality image (in terms of spatial and spectral features) for the improved classification results. This paper uses anisotropic diffusion with PCA for the fusion of SAR and optical data and patch-based SVM Classification with LBP (LBP-PSVM). Fusion results with VV polarization performed better than VH polarization using considered fusion method. For classification, the performance of LBP-PSVM using S1 (VV) with S2, S1 (VH) with S2 is compared with SVM classifier (without patch) and PSVM classifier (with patch), respectively. Classification results suggests that the LBP-PSVM classifier is more effective in comparison to SVM and PSVM classifiers for considered data.

**Keywords**— Fusion, Anisotropic Diffusion, PCA, SVM Classification, LBP

## 1. INTRODUCTION

Image fusion aims to merge the data acquired using different sensors/ resolution to provide an improved image with spectral and spatial features [1]. Image fusion with SAR and optical data is widely used in the field of remote sensing due to their different physical and chemical characteristics. In literature, various image fusion methods were implemented such as dictionary-based fusion, Sparse Representation based fusion, Bayesian fusion, Component Substitution, Multi-resolution Analysis, etc. for the fusion of panchromatic/ multispectral and hyperspectral data [1-2]. Various other methods such as bilateral filter, weighted square filter, Anisotropic diffusion, 3-D Anisotropic diffusion, PCA, Anisotropic diffusion were also introduced to fuse the data and these methods acts as well as edge preserving methods and feature extraction [3]. After fusion, the classification of the fused images is considered to be the necessary step. Classification in remotes sensing involves grouping the pixels having similar characteristics for different land cover types and is widely researched area in the field of remote sensing using satellite imagery. Various classification methods such as Support Vector Machines (SVM) and random forest classifiers are the most researched classifiers by the remote sensing community due to their improved classification results [4]. Local Binary Pattern is also being used for classification purposes. A recent study for hyperspectral dataset reported the higher performance by PSVM in comparison to other classifiers, namely Naïve Bayes, Linear discriminant analysis, K-nearest Neighbors, and Decision Tree classifier [5].

Keeping in view of the effectiveness of Anisotropic diffusion, Anisotropic diffusion with PCA was used to fuse S1 and S2 datasets over an agricultural area in India. In addition to fusion, LBP-PSVM for classification was used to compare the performance with SVM and PSVM to extract both spatial and spectral features of the fused image.

## 2. METHODS USED

This section describes the adopted fusion methods for fusion and classification of S1 and S2 data. For the purpose of fusion, Anisotropic Diffusion with PCA was used and for classification, SVM classifier was used.

### 2.1. Fusion: Anisotropic Diffusion with PCA

Anisotropic Diffusion technique aims to reduce noise (without removing significant features), preserves the edges in an image and smoothen the images at homogeneous distributed regions or edges and preserves the heterogeneous regions with the help of partial differential equations. It is advantageous in the filtering of optical data as it preserves the edges at coarser resolution. It uses flux function which includes the gradient operator and Laplacian operator to preserve the image diffusion of the image [3].

PCA is the linear transform method which is also proposed for fusion and is extremely useful in the image decomposition into low and high frequency components [6]. Anisotropic diffusion with PCA is helpful in performing the fusion to extract the useful spatial features.

### 2.2 Feature Extraction: Local binary pattern (LBP)

LBP is known to be the circular derivative patterns of first-order, obtained by concatenating the binary gradient directions for the image classification. LBP acts as a feature extraction method or a visual descriptor that consists of micropatterns [7]. The histogram of micropatterns generated from LBP contains the image information regarding edge distribution as well as local features. The traditional LBP operator [7] helps in extracting the image information for the constant local grayscale variations. At each pixel location, radius of the neighborhood pixels around the central pixel is calculated to deriving the features.

### 2.3 Classification: SVM

SVM is an optimization technique based on statistical learning which determines the boundary location for each class. Generally, it is designed and considered for the two-class problems which selects the decision boundaries linearly for the separation of the classes [4]. SVM selects a hyperplane for the non- linear separable classes which widens the margin and reduces the misclassification errors simultaneously. A user-defined parameter i.e., regularization parameter (C) acts as trade-off between margin as well as misclassification error. To minimize the excessive computational cost, kernel functions were used in the high dimensionality feature space [4].

## 3. DATASET USED

The study area used in this work is located in Central State farm in Hissar, Haryana (India). Sentinel 1 and Sentinel 2 images of size 964 rows and 1028 columns were acquired 23 March 2019 and 24 March 2019. VV and VH polarized images from Sentine1 and four bands (Red (R), Green (G), Blue (B) and Near Infrared (NIR)) from Sentinel 2 images at 10 m resolution were considered. A total of twelve land cover types, namely, Fallow land, Built-up-area, Dense Vegetation, Fenugreek, Fodder, Gram, Mustard, Oat, Pea, Sparse Vegetation, Spinach and Wheat that were identified after a field visit to the study area on 6 April 2019.

## 4. METHODOLOGY

This section describes the adopted methodology used in this paper for fusion and classification of S1 and S2 data.

For the fusion of both Sentinel 1 and Sentinel 2 data, preprocessed data was considered in this study. 3-D Anisotropic diffusion with PCA was used to fuse Sentinel 1 (VV and VH polarization) and Sentinel 2 data in a band-wise manner. Firstly, the base and details layers from S1 and S2 are extracted using 3-D anisotropic diffusion. Secondly, detail layers are fused using PCA transform and base layers using weighted method. Finally, based layers and detail layers are layer-stacked to form the final fused image.

The quality of the fused images was evaluated using considered fusion indicators, namely, Erreur Relative Globale Adimensionnelle de Synthese (ERGAS), Spectral Angle Mapper (SAM), Relative Average Spectral Error (RASE), Universal Image Quality Index (UIQI), Structural Similarity Index (SSIM), Peak Signal-to-Noise Ratio (PSNR) and Correlation Coefficient (CC).

Once the fused images were obtained, feature extraction on the fused images were performed using pixel-wise Local Binary Pattern for patch-wise SVM classification i.e. (LBP-PSVM). The training and testing of LBP-PSVM classifier, image patch of similar size (i.e., p×p) was extracted with help of ground reference image. During classification, only patches of central pixels whose value is non-zero were assigned a class number in the ground reference image else no class number was assigned. RBF kernel function, optimal values of regularization parameters (C and γ) using a suitable patch size were considered for LBP-PSVM classifier.

The performance of the classification is measured in terms of overall accuracy (OA).

## 5. FUSION AND CLASSIFICATION RESULTS

This section describes the fusion and classification results of S1 and S2 data and the experimental setup was performed using MATLAB and Python 3.0.

Table 1 provides the fusion results of Anisotropic diffusion using PCA in terms of considered fusion indicators for S1 and S2 data. S2 fused with S1(VV) data achieved higher performance in comparison to the S2 fused with (VH) data.

Table 1: Fusion evaluation using various fusion indicators; P (Polarization)

| P | Fusion Parameter using Anisotropic Diffusion with PCA | | | | | | |
|---|---|---|---|---|---|---|---|
|  | ERGAS | SAM | UIQI | SSIM | CC | RASE | PSNR |
| VV | 0.14 | 0.16 | 0.99 | 0.99 | 0.99 | 0.57 | 50.95 |
| VH | 4.13 | 5.69 | 0.91 | 0.98 | 0.98 | 19.76 | 29.46 |

Figure 1 provided the input images and the fusion results for the visually interpretation. The results are provided for both VH and VV polarized images of S1 (Figures 1). The images obtained after fusion with both VH and VV polarization are displayed using three bands (NIR, R, and G). The VV fusion (Figure 1(a)) indicates the better tone, texture of the image and were able to preserve the shape of the sparse vegetated areas.

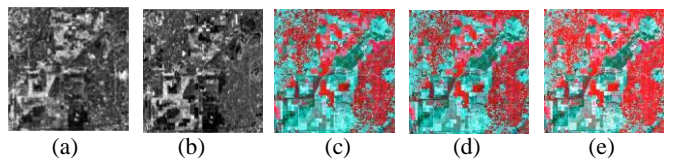

(a)     (b)     (c)     (d)     (e)

Figure 1: Original input images (a) S1 (VV), (b) S1 (VH), (c) S2, Fused image using Anisotropic diffusion with PCA for (d) VV polarization and (e) VH polarization

Figure 1 provided the fusion results for the visually interpretation. The results are provided for both VH and VV polarized images of S1 (Figures 1). The images obtained after fusion with both VH and VV polarization are displayed using three bands (NIR, R, and G). The VV fusion (Figure 1(d)) indicates the better tone, texture of the image and were able to preserve the shape of the sparse vegetated areas.

Table 2 provides the classification results using PSVM, SVM and LBP-PSVM for the obtained fusion results. In Table 3, LBP-SVM shows the better classification results in comparison to SVM and PSVM.

Table 2: Classification Accuracy for SVM, PSVM and LBP-SVM classifiers for obtained fusion image

| Accuracy Measure | Polarization | SVM | PSVM Patch =3 | LBP-PSVM Patch=3 |
|---|---|---|---|---|
| OA | VV | 85.84 | 94.17 | 98.69 |
|  | VH | 85.26 | 93.23 | 97.82 |

Figure 2 provides the classification results for PSVM and LBP-PSVM for VV polarized images only. It can be found from Figure 2 that LBP-PSVM (VV polarization) shows improved performance because it was able to preserve the shape of the vegetation.

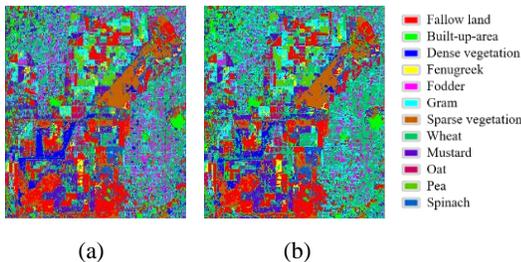

(a)　　　　　　(b)

Figure 2: Classified image for VV polarization using (a) PSVM and (b) LBP-PSVM classifier

## 6. CONCLUSION

This study reports the anisotropic diffusion using PCA to fuse the SAR and optical data. Based on the fusion results, major conclusion drawn from the study is that VV polarized fused images resulted in better performance in comparison to the VH polarized fused images in terms of image analysis for the considered dataset. Another conclusion is that out of PSVM and SVM, only LBP-PSVM were able to improve the accuracy of the classification results using spectral and spatial features acquired from considered patch size. The different classification accuracy of SVM using different approaches suggests that the classifiers are assigning different classes for the same region, therefore more ground data collection is required as well as other deep leaning approaches like Convolutional Neural Networks can be used for the comparison of the classification accuracy.

## 7. REFERENCES

[1] H. Ghassemian, "A review of remote sensing image fusion methods," *Information Fusion*, volume 32, pages 75-89, 2016.
[3] P. Perona and J. Malik, "Scale-space and edge detection using anisotropic diffusion," *IEEE Transactions on Pattern Analysis and Machine Intelligence*, volume 12, no. 7, pages 629–639, 1990.
[2] Kulkarni, C. Samadhan, and P. Rege. Priti, "Pixel Level Fusion Techniques for SAR and Optical Images: A Review," *Information Fusion*, volume 59, pages 13-29, 2020.
[4] G. Mountrakis, J. Im, and C. Ogole, C, "Support vector machines in remote sensing: a review", *ISPRS Journal of Photogrammetry and Remote Sensing*, volume 66, no. 3, pages 247-259, 2011.
[5] D.K. Pathak, and S.K. Kalita, March. "Spectral Spatial Feature Based Classification of Hyperspectral Image Using Support Vector Machine," 6th International Conference on Signal Processing and Integrated Networks (SPIN), pages 430-435, 2019.
[6] Miao Qiguang and Wang Baoshu, "A novel image fusion algorithm using FRIT AND PCA," *10th International Conference on Information Fusion*, pages 1–5, 2007.
[7] T. Ojala, M. Pietikainen, and T. Maenpaa, "Multiresolution gray-scale and rotation invariant texture classification with local binary patterns," *IEEE Transactions on Pattern Analysis and Machine Intelligence*, volume 24, no. 7, pages 971–987, 2002.